\documentclass{article}
\usepackage{ijcai19}

\usepackage{times}
\usepackage{soul}
\usepackage{url}
\usepackage[hidelinks]{hyperref}
\usepackage[utf8]{inputenc}
\usepackage[small]{caption}
\usepackage{graphicx}
\usepackage{amsmath}
\usepackage{booktabs}
\usepackage{algorithm}
\usepackage{algorithmic}
\newcommand\headercell[1]{
   \smash[b]{\begin{tabular}[t]{@{}c@{}} #1 \end{tabular}}}

\title{Deep Bayesian Multi-Target Learning for Recommender Systems}

\author{
Qi Wang$^1$
\and
Zhihui Ji$^1$\and
Huasheng Liu$^1$\And
Binqiang Zhao$^1$
\affiliations
$^1$Alibaba Group
\emails
\{wq140362, jiqi.jzh, fangkong.lhs, binqiang.zhao\}@alibaba-inc.com
}

\begin{document}

\maketitle

\begin{abstract}
With the increasing variety of services that e-commerce platforms provide, criteria for evaluating their success become also increasingly multi-targeting. This work introduces a multi-target optimization framework with Bayesian modeling of the target events, called {\em Deep Bayesian Multi-Target Learning (DBMTL)}. In this framework, target events are modeled as forming a Bayesian network, in which directed links are parameterized by hidden layers, and learned from training samples. The structure of Bayesian network is determined by model selection. We applied the framework to Taobao live-streaming recommendation, to simultaneously optimize (and strike a balance) on targets including click-through-rate, user stay time in live room, purchasing behaviors and interactions. Significant improvement has been observed for the proposed method over other MTL frameworks and the non-MTL model. Our practice shows that with an integrated causality structure, we can effectively make the learning of a target benefit from other targets, creating significant synergy effects that improve all targets. The neural network construction guided by DBMTL fits in with the general probabilistic model connecting features and multiple targets, taking weaker assumption than the other methods discussed in this paper. This theoretical generality brings about practical generalization power over various targets distributions, including sparse targets and continuous-value ones.
\end{abstract}

\section{Introduction}
\begin{figure}
  \centering
    \includegraphics[width=0.5\textwidth]{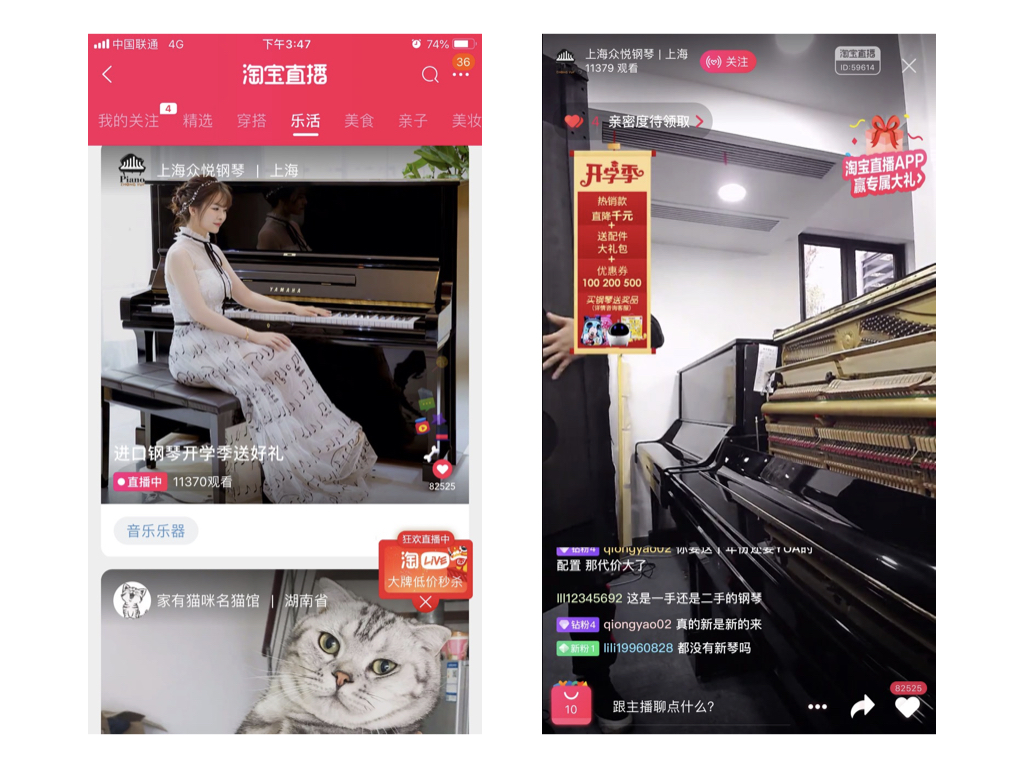}
  \caption{Recommendations display of Taobao live-streaming and layout of a live room}
  \label{figure:live_product}
\end{figure}

Online multi-media platforms usually provide a rich set of interactions with users. This is especially true with the Taobao live-streaming application. As one of the biggest live product promotion platforms on the internet, Taobao live-streaming not only enables users to watch, comment, like and establish connection with live hosts, but also provides various portals towards adding to cart and purchasing behaviors (Figure \ref{figure:live_product}). Therefore it serves the purpose of both an e-commerce platform and a content production/consumption platform. The criteria to evaluate the success of such a multi-media platform are multi-dimensional, concerning not only click-through-rate (CTR) but also many other metrics relevant to user experience such as average user stay time in live rooms, and yet also many other links in the transaction chain. With such a multi-dimensional evaluation system, design of the recommendation system is naturally multi-target, taking many user actions other than the simple click through into the labeled data system.

Multi-target learning has been a very useful line of research in the recommendation system literature. An especially important case is the simultaneous pursuit of click-through-rate (CTR) and conversion rate (CVR) for e-commerce and advertising platforms \cite{ma2018entire,ni2018perceive,chapelle2015simple}. In general, the motivations of doing multi-target learning can be categorized into two perspectives: 1) to balance various performance criteria, especially the conflicting ones; 2) to incorporate auxiliary target information to improve the prediction precision of the primary target. To balance various performance criteria is usually a business requirement. As an example, for Taobao live-streaming the business pursuit includes not only user attention (CTR) but also user experience (user stay time in live room), social connection establishment (follow) and conversion to transactions. The balance between targets is usually achieved by applying weights upon training and inference. 

As a work on learning architectures and their characteristics, discussions in this paper are more relevant to the second perspective, about the potential of incorporating auxiliary target information to improve the primary target. \cite{ruder2017overview} gives a comprehensive discussion on this paradigm. An important observation is that instead of being independent or mutually inhibitive, in many real applications multiple targets are actually highly correlated and possess significant potential of being in synergy. In such cases, gradient descent directions led by different targets can actually guide each other towards a globally better solution, rather than wrestle with each other to reach a mediocre compromise.

Methodology of model design on this line is to try to find a model that can better express the underlying correlation among targets. This can be implemented in deep networks by hard parameter sharing \cite{caruna1993multitask}, soft parameter sharing \cite{evgeniou2005learning,yang2016trace}, and cross-stitch networks \cite{misra2016cross}. From a Bayesian network perspective, the shared layers generate common parents of the targets, and the unshared parameters generate distinct parents. In other words, targets can have common or distinct causes, but they are not causes for other targets. Without changing this big framework, the difference in performance lies in how well the deep structure meets with the real depth of correlation, and how the allocation of shared and distinct parameters fits in with the specific reality of the application.

In practice, implementing common and distinct priors by shared and distinct parameters produces satisfactory results, provided that parameters and the sharing structure are properly tuned. However from a more general perspective, this formulation do not capture the facts that targets can have {\em direct causal effects} on each other. A easy case is recognized in the Taobao live-streaming application where a user cannot have any live room actions and purchasing behaviors until he clicks and enters the live room. Clear-cut causal relationships like this can be hard-coded into the model to regularize its behavior, which is done in ESMM \cite{ma2018entire}. But for less obvious relationships, it is not easy to hard-code.

To handle the direct causal relationships across targets that need to be learned, we propose learning a Bayesian network across target events from data. A lightweight and tempting solution is to estimate the Bayesian network with target nodes only, cutting off the feature side, and then mount back whatever complex feature side deep network to fine-tune. The problem is that when the two networks are estimated separately, it introduces the problem of “explaining-away” \cite{koller2009probabilistic,jensen1996introduction}. In light of the ever more complex deep network architectures on the feature side in today's recommendation systems, and the enormous effect that complex feature information can exert on target relationships, it is unlikely that the separated estimation workflow would produce good results. Causal relationships between targets can be obscure without evidence from features, for example the joint distribution of user stay time in the live room and purchasing behavior may well depend on traits of the live and the user themselves. These are the inspirations and considerations leading to our proposal of an integrated Bayesian framework called {\em Deep Bayesian Multi-Target Learning}, or {\em DBMTL}.

In summary, DBMTL is an integrated feed-forward network modeling the feature-target and target-target relationships simultaneously. As will be shown in Section \ref{sec:theory}, it imposes weaker probabilistic assumption than previous models. We try to model the causal relationships between multiple targets through direct feed-forward MLPs between target nodes, adjusting the direction of each feed-forward link through evaluation and model selection. Letting the model learn cross-inference parameters and directions automatically from data, we avoid making wrong prior assumptions about the causal relationships between target events.

\section{Deep Bayesian Multi-Target Learning}\label{sec:theory}
In this section we introduce the Deep Bayesian Multi-Target Learning framework, or DBMTL. We first formulate the single/multi-target prediction problem in its most abstract probabilistic form. Then we discuss various assumptions adopted by previous models. Finally we describe DBMTL and how it weakens the assumptions. It will then be self-explanatory how DBMTL fits into the spectrum of probabilistic formulations.

\subsection{Probabilistic Formulation for Multi-Target Learning}
In probabilistic form, learning the CTR prediction model in recommendation systems can be formulated as fitting the conditional probability of the click target to the training data. Let $x$ denote the features of an impression, $l$ denote the label whether the impression has been clicked or not. The learning process tries to fit a model $H$ so as to maximize the probability $P(l | x, H)$ . In a feed-forward network setup, $H$ represents the parameters in the multilayer perceptron, mapping the feature end to the predicted target end. Without regularization, this is a maximum likelihood estimation. A regularization on $H$ usually corresponds to certain kind of prior assumption on $H$ so as to make it an MAP estimation, where we instead try to maximize $P(l, H | x) = P(l, | x, H) \cdot P(H)$.

Now if we have two targets to predict, let them be $l, m$ respectively, the formulation now becomes $P(l, m | x, H)$. As an example, in the Taobao live-streaming application, $l$ can represent the binary variable denoting whether a user has clicked and entered a live room, and $m$ can represent the binary variable denoting whether the user has clicked the commodity list button (the commodity list button directs the user to the list of commodities being introduced by the live host, clicking which implies the user's intention of buying something). If more than two targets are concerned, the objective becomes $P(l_1, l_2, l_3, ... | x, H)$.

\subsection{Separation of Target Variables}

\begin{figure}
  \centering
    \includegraphics[width=0.3\textwidth]{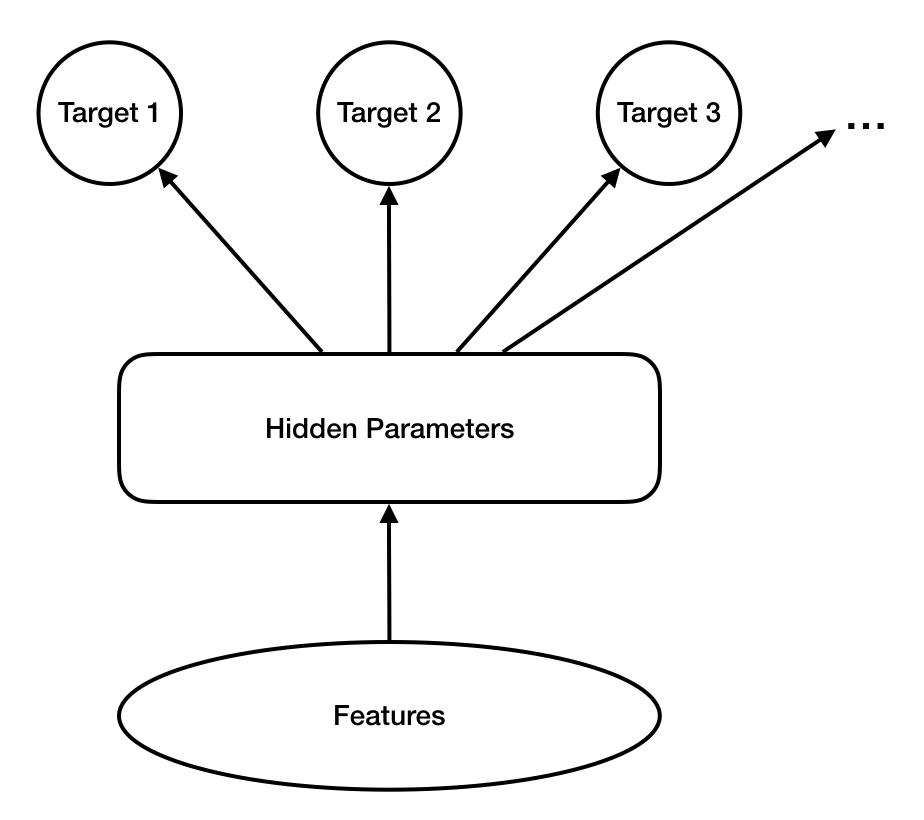}
  \caption{Network Structure with Likelihood Separation}
  \label{figure:hardshare}
\end{figure}

When we have a single binary (or multi-class) target, usually the last stage of prediction is modeled as a logistic regression (or softmax regression) problem. When there are multiple binary targets, we can model them together as a multi-class classification problem in the cartesian space of each target space. But when the number of targets is considerable, the number of categories in the cartesian space blows up exponentially. Each instance of the combined target values can become very sparse in the data so as to make the prediction performance quickly deteriorate. So usually we avoid this exponential space expansion by separating the joint distribution into smaller joint or individual distributions, with certain assumptions about the probabilistic model. For example if we assume the conditional independence of two target events, we can write
\begin{align}\label{eq:separation}
    P(l, m | x, H) = P(l | x, H) \cdot P(m | x, H)
\end{align}
The loss is then split into two distinct terms, and the dimensionality curse of the label space vanishes. Among deep network models, this formulation corresponds to network constructions using hard-shared layers \cite{caruna1993multitask}, which is termed as {\em vanilla MTL} throughout this paper. Abstractly the network topology assumes the pattern illustrated in Figure \ref{figure:hardshare}. The model can take various forms of networks for its hidden layers, but in the final layer it spurs out independent feed-forward branches towards target heads. Most multi-target models in the literature are of this flavor \cite{caruna1993multitask,evgeniou2005learning,yang2016trace,misra2016cross}.

One important observation of our work is that this strong independence assumption in Eq.(\ref{eq:separation}) is not necessary. Using the Bayesian formula, we can instead express the likelihood in Eq.(\ref{eq:separation}) as 
 
\begin{align}\label{eq:dbmtl}
    P(l, m | x, H) = P(l | x, H) \cdot P(m | l, x, H) 
\end{align}

\begin{figure}
  \centering
    \includegraphics[width=0.3\textwidth]{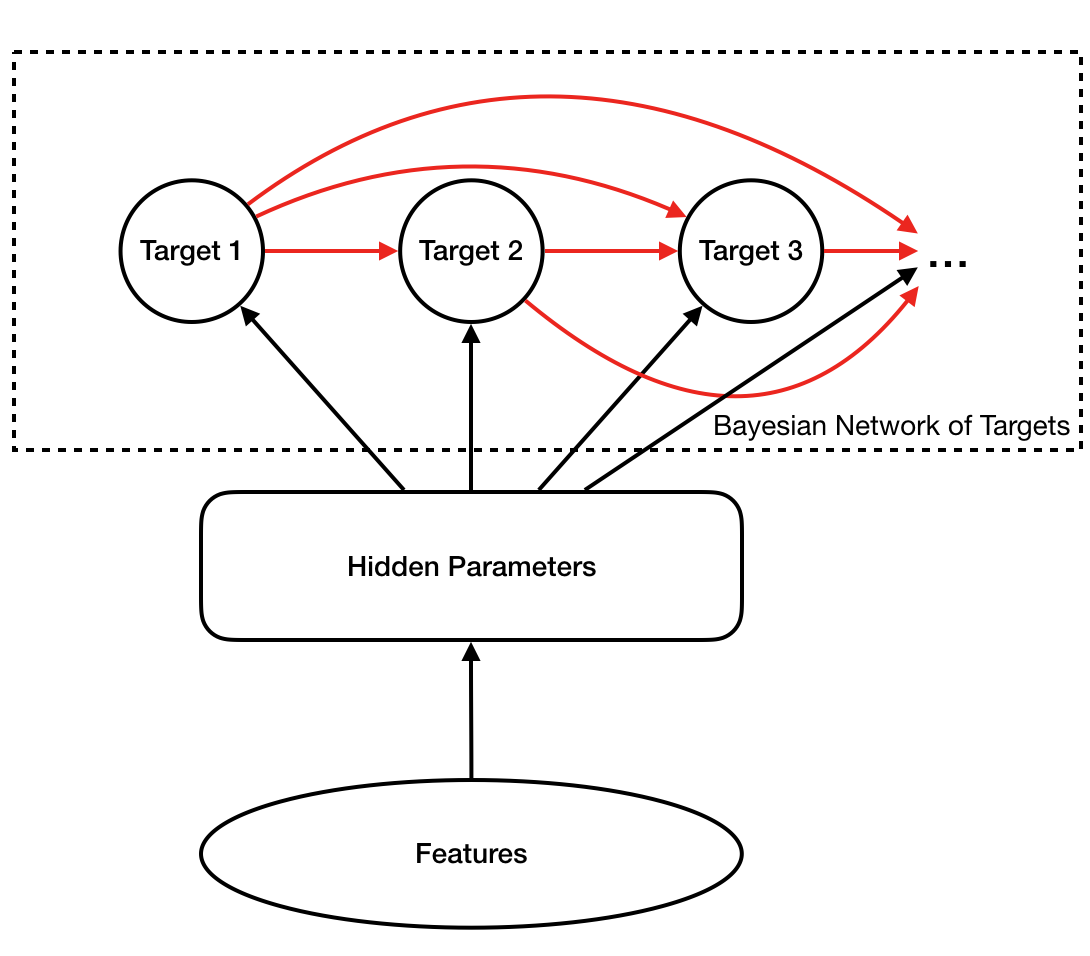}
  \caption{Network Structure for DBMTL}
\label{figure:abstractbayesmtl}
\end{figure}

The equation holds without any assumption. To implement the equation we'll still need to assume that $P(m | l, x, H)$ can be effectively learned, but in general the formulation of Eq.(\ref{eq:dbmtl}) takes much weaker assumption. The corresponding network structure is illustrated in Figure \ref{figure:abstractbayesmtl}.

This is the formulation that we term as DBMTL. We would like to note that the ESMM model \cite{ma2018entire} can be regarded as an instance of such formulation, where $m, l$ must be binary and $P(m=1 | l, x, H)$ is further separated (with explicit assumption about the causality relationship between $l$ and $m$) to the form of $f(x, H) \cdot P(l=1 | x, H)$, where $f(x, H) = P(m^\prime=1 | x, H)$ is an inference function for a virtual binary event $m^\prime$. In our construction, we directly model $P(m | l, x, H)$ as another level of MLP and learn the parameters automatically from data. Compared with ESMM, this theoretically reduces the assumption of $P(m | l, x, H)$ to the entire function space that can be expressed by the MLP. 

Making the cross-target relationship learnable has greater importance when the causal directions between target events is unclear. In such scenarios both directions ($P(l | x, H) \cdot P(m | l, x, H)$ versus $P(m | x, H) \cdot P(l | m, x, H)$) can be trained and tested, and then model selection can determine which is better. It’s not a matter of which construction is “correct” - from the Bayesian perspective, both are correct - it’s a matter of which one is more “learnable” from data. 

All discussions in this section naturally generalize to scenarios of more than two targets, i.e. $P(l_1, l_2, l_3, … | x, H)$. In the Taobao live-streaming application, the target events include user click, live room purchasing behaviors, users' time of stay in a session, interactions in the live room, establishing follow relationships, and many others. 
Most of these target events are binary, while a little subtlety arises when users' time of stay, which is a real-value variable, enters the target set. In such case $p(l_1, l_2, l_3, ... | x, H)$ should be understood as a probability {\em density} rather than a probability, while all derivations and conclusions in this section still hold.

\subsection{Choice of Network Structures}\label{sec:designprinciple}
When there are many targets to be predicted and the causal relationship is obscure, it is generally not feasible to iterate over all setups and compare their results. The number of relationships between targets is $O(n^2)$ and the number of all possible Bayesian network setups is $2^{O(n^2)}$. Simple techniques for Bayesian network structure learning can be used to reduce the space of exploration \cite{koller2009probabilistic}. One way is to build up the Bayesian network incrementally and greedily, i.e. adding target nodes incrementally, fixing all existing edges unchanged, and only iterating through links relevant to the new node to determine best directions. This technique reduces the complexity to $O(n^2)$. Another way is to simply provide an initial network structure based on intuition and prior knowledge about the causal relationships, then apply local variations within certain limit of iterations, keeping good variations and drop bad ones. Some principles can guide us towards a relatively good initial design. Directions with natural causal relationship is usually better than directions with natural anti-causal relationship. And it is generally better to use a more evenly distributed target to predict a less evenly distributed target, than the reverse. For example “Follow” button click rate is less evenly distributed than “Goods Bag” button click rate (follow button click is relatively rare while the ratio of live room users that will click the goods bag button is relatively closer to 1/2), therefore we can expect worse result using the follow event as the cause than using goods bag click as the cause, which is confirmed in our experiments. These principles are to be made clearer in Section \ref{section:experiment}.

\section{Implementation of DBMTL and Experiments}\label{section:experiment}
In this section we give a detailed description of our real implementation of the DBMTL network. Then we run experiments to demonstrate traits of the model from several perspectives.

\subsection{The DBMTL Network Structure}
\begin{figure*}
  \centering
    \includegraphics[width=0.8\textwidth]{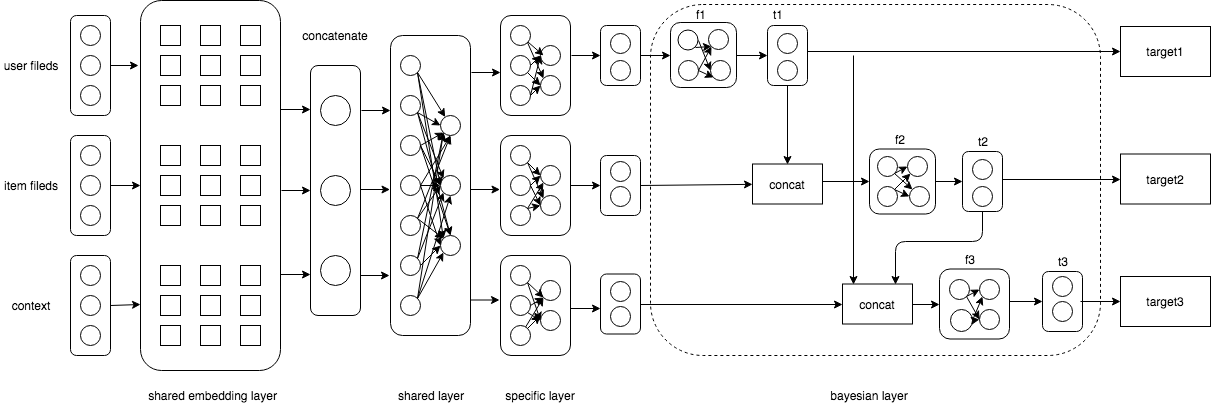}
  \caption{Concrete Structure of DBMTL Deep Network}
  \label{figure:concretedbmtl}
\end{figure*}

Our implemented DBMTL framework (Figure \ref{figure:concretedbmtl}) includes input layer, shared embedding layer, shared layer, specific layer and Bayesian layer. Shared embedding layer is a shared lookup table,  where shared embedding features are learned across different targets. Shared layer and specific layer are multilayer perception (MLP), which captures the common features and specific features from different targets respectively. Bayesian layer is the most import part in DBMTL. As for the instance shown in Figure \ref{figure:concretedbmtl}, it implements the Bayesian formula

\begin{align}
&P(t_1,t_2, t_3 | x,H) = \nonumber \\
&\quad P(t_1 | x, H) \cdot P(t_2 | t_1, x, H) \cdot P(t_3 | t_1, t_2, x, H) 
\end{align}

The corresponding negative log-likelihood loss is 
\begin{align}
L(x, H) =& -\log(P(t_1,t_2, t_3 | x,H)) \nonumber \\
=&-(\log(P(t_1 | x, H)) + \log(P(t_2 | t_1, x, H))\nonumber \\
&+ \log(P(t_3 | t_1, t_2, x, H)))
\end{align}

For practical reasons, different weights is applied to each term to control the relative importance of various targets, transforming the loss to a form of
\begin{align}\label{eq:objective}
L(x, H) =& -\log(P(t_1,t_2, t_3 | x,H)) \nonumber \\
=&-(w_1\log(f_1(x, H))+ w_2\log(f_2(t_1, x, H))\nonumber\\
&+w_3\log(f_3(t_1, t_2, x, H)))
\end{align}

where $f_1^{w_1}(x, H) = P(t_1 | x, H)$, $f_2^{w_2}(t_1, x, H) = P(t_2 | t_1, x, H)$ and $f_3^{w_3}(t_1, t_2, x, H) = P(t_3 | t_1, t_2, x, H)$.

Functions $f_1, f_2, f_3$ in the Bayesian layer are implemented as fully connected perceptrons or MLP to learn the hidden relationship among targets. Each concatenates the embeddings of its inputs as the MLP input and outputs an embedding of the output target. Each target embedding then goes through a final linear-logistic layer to generate the final probability of target.

In our experiment, DBMTL is implemented with Wide\&Deep Framework of Tensorflow \cite{cheng2016wide}. The model parameters are learned via minimizing the objective function in Eq.(\ref{eq:objective}), and Adaptive Moment Estimation(Adam) \cite{kingma2014adam} is adopted for faster convergence with a batch size of 2000 and learning rate of 0.001. To avoid overfitting, L1 Norm, L2 Norm and Dropout \cite{srivastava2014dropout} techniques are used on network layers. With about 1 billion training samples, hyperparameters are tuned as follow: 64 for embedding size, [256, 128, 64] for shared layer, [64, 32] for specific layer, [32] for Bayesian layer.

\subsection{Experimental Setup}
In the Taobao live-streaming application, users can interact in multiple ways, and there are correspondingly many dimensions to evaluate its success. We give a term for each as below:

\begin{itemize}
\item \textbf{Click Through Rate (CTR)} – the percentage of impressions that result in click and entrance of the live room.
\item \textbf{Goodslist Conversion Rate (CGR)} – in a live room, there is a “Goods Bag” button where users can click to view the list of goods being introduced (users can then select a good of interest and be forwarded to the purchase page). The Goodslist Conversion Rate is defined as the percentage of live room users that have clicked the goods bag button.
\item \textbf{Follow Conversion Rate (CFR)} – a user can follow a live host so that when the host starts living, the user gets notified. The Follow Conversion Rate is defined as the percentage of live room users that have resulted in following behavior.
\item \textbf{Comment Conversion Rate (CCR)} – a user can send real-time comment in a live room. The Comment Conversion Rate is defined as the percentage of live room users that have comment behavior.
\item \textbf{Like Conversion Rate (CLR)} – a user can “like” the live host in a live room. The Like Conversion Rate is defined as the percentage of live room users that have clicked the “Like” button.
\item \textbf{Average Stay Time (AST)} – The time a user spends before leaving the live room is an important indication of the user’s interest and his satisfaction of the content of the live. The Average Stay Time is defined as the average time that users spend in live rooms.
\end{itemize}

According to these evaluation dimensions, loss heads can be easily constructed for each target. For CTR, CGR, CFR, CCR and CLR, because they correspond to binary outputs, they are associated with logistic loss heads. For AST, it is a real-value regression problem, and mean square error (MSE) is adopted as loss function. In consideration of the scale problems of our real data, we use logarithm of stay time, instead of its original value as the AST label.

The dataset for our experiments comes from online logs. In a certain time window, data from the first 15 days are taken as training data, and samples of the next day are used to test the performance. Over a hundred features are extracted from each data sample, including many large-scale sparse id features. For the label part, since all live room interactions are dependent on user entering the room in the first place, the CGR, CFR, CCR, CLR and AST labels are only turned “on” (assigning value 1 for binaries, assigning the value as it is for AST) when the CTR label is “on”. Otherwise an “off” (value 0 for binaries, 0.0 for AST) is assigned. In all experiments demonstrated below, the same feature extraction workflow is used for each method tested.

\subsection{General Performance on Multiple Targets}
In the first experiment, prediction performance on the 6 targets is evaluated and compared among various learning methods. In this experiment, the structure among targets is designed as the CTR target pointing to the others, which follows the natural causality principle and the weights of CTR, CGR, CFR, CCR, CLR, AST losses are set as [0.7, 0.05, 0.05, 0.0, 0.0, 0.1]. Note that this weight setting on the one hand reflects the business view about the importance of each product target, e.g. user stay time is a relatively more important auxiliary target than goods bag click or follow; Also note that CCR and CLR targets are intentionally left as 0.0 to test the generative power of the model towards no-training targets (to be made clearer in the analysis of results). The methods tested include the single-task Wide\&Deep network \cite{cheng2016wide}, vanilla MTL \cite{caruna1993multitask}, ESMM \cite{ma2018entire} and our DBMTL framework. AUC for each binary target and MSE for the continuous target are evaluated, as is displayed in Table \ref{tab:generalperformance}.

From the results we can observe that: 1) Multi-target learning models in general achieve better performance than the single-target model. Despite that single-target model is specifically tuned to optimize the CTR target, Multi-target models can excel on the primary target. We believe this is good evidence in support of adding auxiliary targets to improve the main target, even when auxiliary targets performance are not actually cared about \cite{ruder2017overview}. The concept manifests especially well in our application, and we believe it’s because auxiliary behaviors in Taobao live-streaming (goods bag click, follow, comment, like and stay time in the live room) are very strong indicators of the quality of the live, which in turn is a valuable information source to predict the CTR target. 2) Among all three multi-target learning models, DBMTL has significantly better performance on all the 6 targets. This supports the analysis in Section \ref{sec:theory} that with Bayesian network modeling across targets, inter-target causality relationship can be better captured, and the prediction performance benefits from weaker statistical assumption. Note also, the effect manifests well in our scenario possibly because the targets concerned have intricate black-box causal relationships that is better expressed by a MLP. 3) Despite that we have intentionally left the CCR and CLR targets un-trained, ESMM and DBMTL can learn significant information about these two targets (vanilla MTL does not have this ability since it does not model cross-target relationships). This on the one hand reinforces the merit of using auxiliary targets to enhance the concerned targets, on the other hand it further implies that inter-target Bayesian modeling can indeed benefit learning.

\subsection{Performance on Target Pairs}
In this experiment, we take a closer look at the performance of MTL methods on auxiliary targets with different traits. Specifically, we select CGR – a binary target with normal sparsity (a large percentage of users in the live room will click the goods bag button), CFR – a binary target with extensive sparsity (a small percentage of users in the live room will click the follow button), and AST – a real-value target. Each auxiliary target is combined with CTR to form an optimization pair. Performance on each optimization pair is listed in Tables \ref{tab:ctr-cgr}, 
\ref{tab:ctr-cfr}, \ref{tab:ctr-ast}, respectively.

\begin{table}
\centering
\begin{tabular}{@{} *{7}{c} @{}}
\toprule
\headercell{Learning\\Models} & \multicolumn{6}{c@{}}{Targets}\\
\cmidrule(l){2-7}
& CTR &  CGR & CFR & CCR & CLR & AST   \\   
\midrule
WDL     & 0.7008  & 0.5 &	0.5 &	 0.5 & 0.5 & 	0.1316 \\
vanilla MTL &	0.7139 &	0.7456 &	0.7298 &	0.5 &	 0.5 &	0.1297 \\
ESMM &	0.7113 &	0.7462 &	0.7206 &	0.5726 &	0.5596 &	0.1246 \\
DBMTL &	\textbf{0.7159} &	\textbf{0.7535} &	\textbf{0.7355} & \textbf{0.7117}	 & \textbf{0.5702} & 	\textbf{0.1206} \\
\bottomrule
\end{tabular}
\caption{Performance of various models on the 6 targets (AUC for binaries and MSE for AST)}
\label{tab:generalperformance}
\end{table}

\begin{table}
\centering
\begin{tabular}{@{} *{3}{c} @{}}
\toprule
\headercell{Learning\\Model} & \multicolumn{2}{c@{}}{Targets}\\
\cmidrule(l){2-3}
& CTR &  CGR   \\   
\midrule
vanilla MTL &	0.7101& 	0.7478 \\
ESMM &	0.7105 &	0.7500 \\
DBMTL &	\textbf{0.7163} &	\textbf{0.7568} \\
\bottomrule
\end{tabular}
\caption{Target Pair Performance: CTR-CGR (AUC)}
\label{tab:ctr-cgr}
\end{table}

\begin{table}
\centering
\begin{tabular}{@{} *{3}{c} @{}}
\toprule
\headercell{Learning\\Model} & \multicolumn{2}{c@{}}{Targets}\\
\cmidrule(l){2-3}
& CTR &  CFR   \\   
\midrule
vanilla MTL &	0.7129 &	0.7260 \\
ESMM &	0.7108 &	0.7328\\
DBMTL &	\textbf{0.7157} &	\textbf{0.7351} \\
\bottomrule
\end{tabular}
\caption{Target Pair Performance: CTR-CFR (AUC)}
\label{tab:ctr-cfr}
\end{table}

\begin{table}
\centering
\begin{tabular}{@{} *{3}{c} @{}}
\toprule
\headercell{Learning\\Model} & \multicolumn{2}{c@{}}{Targets}\\
\cmidrule(l){2-3}
& CTR &  AST   \\   
\midrule
vanilla MTL &	0.7105 &	0.1308\\
ESMM &	0.7095 &	0.1287 \\
DBMTL &	\textbf{0.7178} &	\textbf{0.1219} \\
\bottomrule
\end{tabular}
\caption{Target Pair Performance: CTR-AST (AUC for CTR and MSE for AST}
\label{tab:ctr-ast}
\end{table}

The observations from this experiment are: 1) DBMTL outperforms other methods in all three experiments, demonstrating its generality among various target types. 2) ESMM and DBMTL are both more successful in modeling the auxiliary target, while the improvement is not as significant in the CTR-CGR (non-sparse) case as in the CTR-CFR (sparse) case. Referring to the analysis of \cite{ma2018entire}, this may be due to that inter-connection of targets brings especial gain when learning sparse targets, since the sparse target can take advantage of information from non-sparse primary target data. 3) The improvement of DBMTL over ESMM is more significantly seen in the CTR-AST (real-value) case, showing the generalization power of DBMTL when dealing with continuous-value targets. 

\subsection{Varying Bayesian Structures}
The Bayesian network structure, i.e. directions of connections in the acyclic Bayesian network can have significant influence on the performance. To demonstrate the design principles we propose in Section \ref{sec:designprinciple}, we select a group of three targets and three structures, each assuming the structure of one target event pointing towards the other two target events. Comparisons of their performances are shown in table \ref{tab:performancebayesianstructure}.

\begin{table}
\centering
\begin{tabular}{@{} *{6}{c} @{}}
\toprule
\headercell{Bayesian\\Structure} & \multicolumn{3}{c@{}}{Targets}\\
\cmidrule(l){2-4}
& CTR & CGR & CFR  \\   
\midrule
CTR $\rightarrow$ others &	\textbf{0.7159} &	\textbf{0.7535} &	\textbf{0.7355}\\
CGR $\rightarrow$ others &	0.7127 &	0.7476 &	0.7289\\
CFR $\rightarrow$ others &	0.7080 &	0.7431 &	0.7102\\
\bottomrule
\end{tabular}
\caption{Performances of Different Bayesian Structures (AUC for binaries and MSE for AST)}
\label{tab:performancebayesianstructure}
\end{table}

Consistent with our intuitive guess, the natural causal direction CTR$\rightarrow$others (natural in the sense that users can only have other behaviors once clicked and entered the live room) yields the best score in all criteria. We believe the reason is that “correct” causal relationships can in general be more efficiently modeled and learned. CGR$\rightarrow$others direction scores better than the CFR$\rightarrow$others direction in all criteria, making evident the other design principle, that we should favor evenly distributed targets pointing towards unevenly distributed targets rather than the reverse (the ratio of positive versus negative samples for the CGR target is much more evenly distributed than for the CFR target). The rationality is that evenly distributed targets contain more information (has a higher entropy value) than unevenly distributed targets.

\subsection{Performance in Online Taobao live-streaming Environment}
The online deployment of DBMTL has brought significant improvement to the Taobao live-streaming application. DBMTL improves the online CTR, CGR, CFR, CLR, CCR, AST by 4.41\%, 3.06\%, 2.91\%, 10.23\%, 5.95\%, 4.99\% respectively relative to vanilla MTL (ESMM performs not as well as vanilla MTL). The improvement is measured in an online A/B test during 2 weeks. In the online deployment the weights of CTR, CGR, CFR, CCR, CLR, AST are set as [0.7, 0.05, 0.05, 0.05, 0.05, 0.1] for training as well as predicition. Note that performances and appropriate parameters may well vary according to specific applications.

\section{Conclusions}
For the multi-target learning problem, we propose the DBMTL formulation, modeling the causal relationships among targets explicitly using a Bayesian network structure across target heads. DBMTL outperforms single-target WDL and other MTL methods on the Taobao live-streaming dataset. The success of DBMTL lies in its integral way of modeling the causal relationship among targets and features, weakening many assumptions that other deep MTL structures adopt for the underlying probability model. Since DBMTL framework does not make specific assumption about target distributions and types, it readily generalizes to various distributions and value types. We also propose two principles in designing the Bayesian structure: respecting clear-cut natural causalities, and favoring more-entropy targets pointing to less-entropy targets.

Apart from the business merits the targets themselves behold, multi-target learning in the merit of taking advantage of auxiliary targets to enhance the primary target can also be regarded as a step towards blurring between features and labels, or more generative rather than discriminative models. However efficient learning of the Bayesian structure is still a challenging task when the number of nodes becomes high, which is a major obstacle towards the “more generative” direction.

\section*{Acknowledgments}
We owe special thanks to the Taobao live-streaming product and engineering team who helped deploy our recommendation system. Engineering team of the Taobao Personalization Platform also provide precious assistance and advisory in deployment of the system. We thank members of the Alibaba Recommendation Algorithm Team for various fruitful discussions that contributed to the evolution of the Taobao live-streaming recommendation system.

\bibliographystyle{named}
\bibliography{ijcai19}

\end{document}